# Detection of Late Blight Disease in Tomato Leaf Using Image Processing Techniques

Muhammad Shoaib Farooq[1], Tabir Arif[1], Shamyla Riaz[1]
[1]Department of Artificial Intelligence, School of System and Technology, University of Management and Technology, Lahore, 54000, Pakistan
Corresponding author: Muhammad Shoaib Farooq (shoaib.farooq@umt.edu.pk)

**ABSTRACT** One of the most frequently farmed crops is the tomato crop. Late blight is the most prevalent tomato disease in the world, and often causes a significant reduction in the production of tomato crops. The importance of tomatoes as an agricultural product necessitates early detection of late blight. It is produced by the fungus Phytophthora. The earliest signs of late blight on tomatoes are unevenly formed, water-soaked lesions on the leaves located on the plant canopy's younger leave White cottony growth may appear in humid environments evident on the undersides of the leaves that have been impacted. Lesions increase as the disease proceeds, turning the leaves brown to shrivel up and die. Using picture segmentation and the Multi-class SVM technique, late blight disorder is discovered in this work. Image segmentation is employed for separating damaged areas on leaves, and the Multi-class SVM method is used for reliable disease categorization. 30 reputable studies were chosen from a total of 2770 recognized papers. The primary goal of this study is to compile cutting-edge research that identifies current research trends, problems, and prospects for late blight detection. It also looks at current approaches for applying image processing to diagnose and detect late blight. A suggested taxonomy for late blight detection has also been provided. In the same way, a model for the development of the solutions to problems is also presented. Finally, the research gaps have been presented in terms of open issues for the provision of future directions in image processing for the researchers.

## I. INTRODUCTION

One of the most consumed crops worldwide is tomatoes. The most prevalent tomato disease is late blight, which is caused by Phytophthora, a fungus-like organism that prefers humid environments with a certain temperature range. Within a few days of the first lesion appearing on the farm, this disease can damage the entire field's crops, resulting in significant yield loss [1].

As a result, early diagnosis of this illness is critical to avoiding massive economic damages. This fungus is transmitted through the air by branching hyphae that form from the stomata of diseased plants. Clouds love it because it protects the bacteria from UV light and damp conditions, letting that contaminate plants when they land [2].

Therefore, lesions on leaves may emerge within three to five days. Water-soaked lesions with white monogenesis on the border, commonly appearing on the bottom side and infrequently on the top side of the leaf and appearing beige in color after lesions dry out in hot weather, are the most common signs of this illness [3].

Identifying plant disease manually with the bare eye requires a team of experts and regular monitoring. It is expensive to run a huge farm. Therefore, image processing methods might be used to identify diseases in leaves automatically, saving time, money, and effort over previous approaches. Crop output is enhanced by the early diagnosis of diseases in leaves. Using image processing techniques such as segmentation, detection, and classification, disease-affected leaves may be identified early on, and crop production and quality can be increased [4].

When plant sickness is discovered via leaf symptoms, the approach is more accessible and less expensive. Using an automated detection approach saves time, labor, and precision. Identifying leaf disease manually with the bare eye requires a team of experts and regular monitoring. It is expensive to run a huge farm. Consequently, image processing techniques might be used to identify diseases in leaves automatically, saving time, money, and effort over previous approaches [5]. Crop output is increased by the early identification of diseases in leaves.

The goal of this study is to give a thorough, systematic literature evaluation on the methods used to detect late blight in image processing. The following are the domain-related contributions made by this paper:

The following is how this document is structured: The research methodology used in this review is described in Section II, which also includes the study objectives, research questions, and motivations. It also includes the search strategy, article selection process, abstract-based keywording, and quality evaluation criteria. To consider the extracted findings and methodically respond to the study questions, analysis and results representation have been offered in section III. Section IV provides thorough discussions, a proposed taxonomy, and a general framework for a late blight prediction model to synthesize the study findings. In section V, unresolved problems and difficulties have been highlighted to provide the research community with impending viewpoints. Finally, section VI has completed the review.

## II. RELATED WORK:

Various researchers have attempted to use image processing techniques to detect illnesses that create symptoms such as spots, patches, lesions, and variations in surface roughness. Anthracnose disease affects mango, grapes, and pomegranates in general. The lesion zones were divided and graded according to the percentage of the affected area. Image processing can be used for automatic plant disease recognition and diagnosis system. [6]. They used a neural network classifier to distinguish between apples with and without anthracnose. Normal and impacted anthracnose fruit types, on the other hand, had classification accuracies of 84.65% and 76.6 percent, respectively.

Similar research was conducted [7] on late blight disease identification in tomato plantations using distant sensing spectral images. Only in the third stage of the sickness could it be diagnosed, not in the early stages. To identify between healthy and unhealthy plants, the author employed a 5-index feature vector technique. To categorize the illness, [8] separated sick plant leaf areas and extracted their textural characteristics. With an overall accuracy of 87.66%, support vector machines were used to categorize leaves from banana, bean, guava, jackfruit, lemon, mango, potato, and tomato for various illnesses. [9] suggested a software-based system for automated disease identification and categorization in plants. Their approach is based on a color transformation framework for the raw leaf picture, rather than image augmentation techniques. After applying color space transformation, the image was segmented using K-means clustering. Segmented contaminated pieces were used to gather texture attributes. The gathered traits were sent into a neural network to classify the leaves. A method for identifying nitrogen and potassium shortages in tomato plants was proposed [10]. From the color picture, the program retrieved several attributes. Difference operators, Fourier transform, and Wavelet packet decomposition were also used to extract texture characteristics. A genetic algorithm was used to determine which traits should be included. This diagnostic method had an accuracy of more than 82.5 percent, and it could identify illnesses 6–10 days before specialists could. A Fuzzy logic-based technique was used for quantifying illness symptoms. Pomegranate leaves served as the test photos. The algorithm begins the process of transforming the photos into a color space. Through K-means clustering, the pixels are divided into several classes. One of the groups will correlate to the afflicted regions, but the text does not explain how to determine which group is accurate. The proportion of the leaf that is infected is determined in the following steps. Finally, for the final estimation of the illness grade, a Fuzzy Inference System is used.

Various researchers have attempted to use image processing techniques to detect illnesses that create symptoms such as spots, patches, lesions, and variations in surface roughness. Anthracnose disease affects mango, grapes, and pomegranates in general. By [11], the lesion zones were divided and graded according to the percentage of the affected area. Using a neural network classifier, they distinguished between apples with normal and anthracnose infection. Normal and impacted anthracnose fruit types, on the other hand, had classification accuracies of 84.65% and 76.6 percent, respectively. [7] conducted similar work on late blight disease identification in tomato plantations using distant sensing spectral images.

Even though late blight appears to be successful and appropriate in various literature studies [10], the research has considerable limitations. Most studies that have been done so far have reviewed how to identify late blight during training. Furthermore, models that have been proposed thus far are constrained, and they are environment-dependent. It may result in the capture of various behaviors that make the categorization and prediction of diseases using image analysis more challenging. In addition, the reviewers failed to offer any model or taxonomy that could have aided the researchers and developers.

However, this study presented a systematic review of the recent research conducted on late blight disease detection solutions to cope with the problems related to late blight in plants. The novelty of this work is that I have presented a taxonomy of elements that will help farmers in the future. Lastly, open issues and directions for future research have been identified

## III. RESEARCH METHODOLOGY

The SLR lays forth a step-by-step process for gathering, investigating, and determining main articles from all accessible studies in the topic under evaluation. For impartial data gathering and demonstration of analyzed and extracted outcomes, the standards for SLR proposed were followed in this work.

The research method for this SLR is depicted in Figure 1. There are six levels to this suitable and reflective review procedure: 1) Research objectives must be defined. 2) Research questions are defined. 3) deciding on a search method 4) Paper screening and selection 5) Keywords are used to categorize documents. 6) extraction and synthesis of data

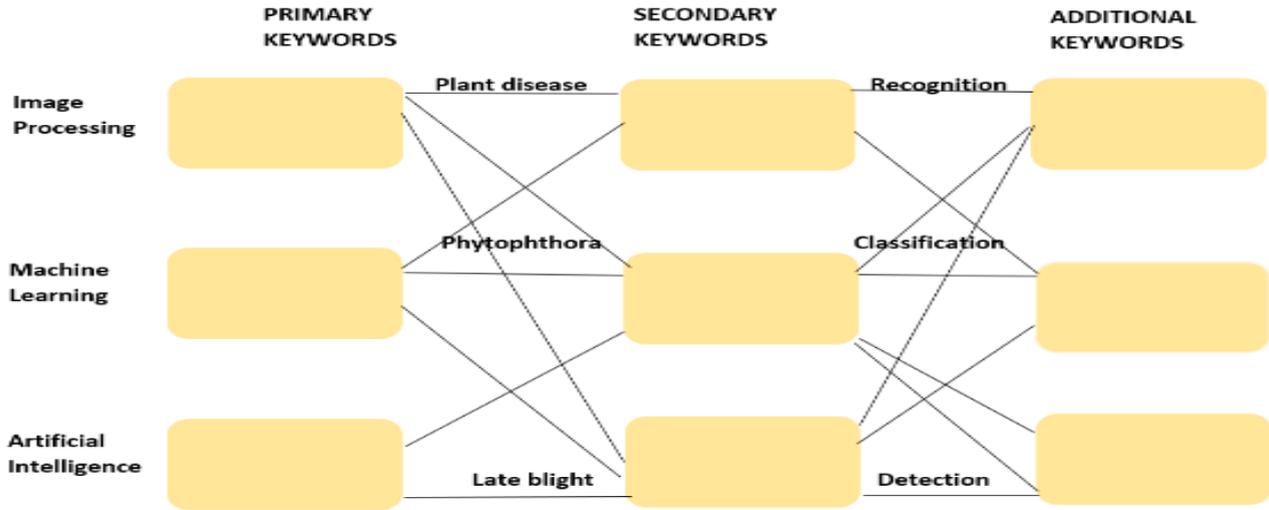

*Figure 1. Terms and keywords used in search*

### A. RESEARCH OBJECTIVES (RO)
The main intents of this study are:
RO1: To determine the severity of tomato Late Blight disease using image processing techniques
RO2: To give a detailed review and analysis of the work.
RO3: To determine which technique is most

### B. RESEARCH QUESTIONS (RQ)

To appropriately conduct this SLR, the major research topics have been identified. Additionally, as part of the study, a thorough search strategy for locating and extracting the most significant articles was developed. The research questions discussed in this study are listed in Table I along with their primary goals. The questions are addressed and answered in accordance with the procedures outlined in [26], [27].

**TABLE I. RQ and major motivations**

| | Research Question | Major Motivation |
|---|---|---|
| RQ1 | What types of image processing models and techniques have been used to improve the quality of tomato crops affected with late blight disease? | To identify the importance of image processing in the field of architecture |
| RQ2 | What type of challenges and opportunities existing image processing for the diagnosis of late blight? | This question intends at recognizing the strength and limitations of current image processing models |
| RQ3 | What types of datasets are available for the detection of late blight? | This study examines the accessibility of benchmark datasets as well as internet-collected datasets that are not published or listed. |

### C. SEARCH SCHEME

The formulation of a search plan to effectively identify and gather potentially significant articles in the targeted field is a critical element of SLR. This process includes the description of a search string, the selection of the literature sources for the search, and the segregation (inclusion/exclusion) strategy used to identify the most pertinent articles in the collection. Several characteristics of the collected publications were evaluated qualitatively and empirically to represent distinct study views.

1) SEARCH STRING
A successful and impartial investigation was achieved by using a keyword-based string to search and gather available articles on the topic using several well-known digital research libraries. To ensure the validity of the search string in terms of the relevance of its findings, the key ideas have been examined in light of the research questions in order to obtain relevant keywords and terms utilized in the chosen field of study. The completed keywords and their alternative terms required to create a search string for locating the most significant articles are listed in Figure 1. To generate a search string, the logical "AND" and "OR" operators were employed to combine the completed keywords and alternative terms. When zero or more characters were necessary, the wild character "*" was also used. The "OR" operator allows for more search possibilities, but the "AND" operator concatenates the phrases to specify search alternatives and limits the query to get appropriate search results.

The final search string has three parts. The first part of the string is used to limit findings associated with computer-based or computational research, the second part is used to limit results associated with late blight prediction, and the last element is used to limit results associated with non-computational studies.

The search phrases from Table II are coupled with the "for all," "for the OR," and "for the AND" operators in (1) to formalize the whole search string for each chosen repository. R stands for search results received in contradiction of a search string. (1) can be stated as a general search term: (("digital" "image processing" OR artificial intelligence OR machine learning) AND ("late blight detection" OR "detection of late blight" OR "phytophthora recognition")

2) LITERATURE RESOURCES

To carry out a literature search using online databases for publishing and gathering research, field specialized and most prominent journals were chosen. Table II lists the details of the search strings used and the results for the selected repositories.

**TABLE II. Publisher wise search strings**

| Repository | Search Strings |
|---|---|
| Scopus | TITLE-ABS-KEY ("IMAGE PROCESSSING" OR "MACHINE LEARNING" OR ARTIFICAL INTELLEGENCE") AND("CIRCULAR LESION", OR "LATEBLIGHT",OR"PYTOPHTHORA")AND(RECOGNITON"OR"CLASSIFICATION" OR "DETECTION |
| Wiley online | IMAGE PROCESSING OR MACHINE LEARNING OR ARTIFICAL INTELLIGENCF AND LATE BLIGHT OR CIRCULAR LESIONS OR PYTOPHOTHORA |
| Springer Link | (DIGITAL IMAGE PROCESSING" OR "MACHINE LEARNING" OR "NEURAL NETWORK" OR "ARTIFICIAL INTELLIGENCE") AND ("CIRCULAR LESION", OR "LATE BLIGHT",OR"PYTOPHTHORA")AND ("RECOGNITON"OR"CLASSIFICATION"OR "DETECTION")) |
| Science Direct | ("DIGITAL IMAGE PROCESSING" OR "MACHINE LEARNING" OR "NEURAL NETWORK" OR "ARTIFICIAL INTELLIGENCE") AND("CIRCULAR LESION", OR "LATE BLIGHT",OR"PYTOPHTHORA")AND ("RECOGNITON"OR"CLASSIFICATION"OR "DETECTION") |
| IEEE Xplore | ((" DOCUMENT TITLE: "IMAGE PROCESSING" OR "MACHINE LEARNING" OR "ARTIFICIAL INTELLEGENCE"AND("ABSTRACT":"CIRCULAR LESIONS"OR"LATE BLIGHT TOMATO")AND("DIAGONSIS"OR "CLASSIFICATION"OR "DETECTON") |
| ACM Digital Library | (("IMAGE PROCESSING" OR "MACHINE LEARNING" OR "ARTIFICIAL INTELLIGENCE" OR "NEURAL NETWORK" OR "VECTOR MACHINE") AND ("ABSTRACT": "TOMATO PLANT PHYTOPHTHORA" OR"LATE BLIGHT OR"CIRCULAR LESIONS")AND("RECOGNITION"OR"CLASSIFICATION "OR "DETECTION")) |

3) INCLUSION AND EXCLUSION CRITERIA
Parameters defined for inclusion criteria (IC) are:
**IC 1)** Include research that was mostly done to anticipate late blight in tomato plants using image processing technology.

**IC 2)** The articles featured recent and developing studies on employing image processing to identify late blight. All papers met the exclusion criteria, which excluded any that did not focus on the binary categorization of late blight disease.
**EC 1)** Articles printed in publications other than technical reports, journals, patents, and conferences
**EC 2)** Articles without a definition of the data sources or with an unclear data collecting process.
**EC 3)** Articles that are not published in English.
**EC 4)** Articles released prior to 2001.
**EC 5)** Articles that are unrelated to the search string

### D. SELECTION OF RELEVANT PAPERS

There is repetition and a significant number of research articles from the primary search strategy, not all of which are directly related to the stated research questions. Therefore, in order to find publications that are actually crucial, the searched papers must be reevaluated and inspected. The studies must be sorted by title and any duplicates must be omitted. Numerous publications that were accessible for this study but unrelated to the chosen subject were vetted based on their titles, and any unsuitable papers were discarded. Finally, by keeping to the specified criteria, the chosen articles were added to the subsequent evaluation level. After applying all of these filters, snowball tracking was used to check that no important study was missing by looking up each study's references. A total of 30 primary studies were included in this review after seven additional studies were chosen at this point and added to the papers that had already been chosen. Table V describes the whole Selection process.

### E. ABSTRACT BASED KEYWORDINGS
To determine the fundamental aspect of the work, its contribution to the topic, and the most relevant keywords, the abstract was originally analyzed. The keywords collected from several articles are then combined, which results in a clear understanding of the study's contribution. The papers were then identified for review processing using these keywords.

### F. QUALITY ASSESSMENT CRITERIA

In an SLR, a quality assessment (QA) is typically conducted to rate the quality of chosen articles. A questionnaire has been created for this SLR to assess the caliber of the chosen articles. The prior mapping research is used to guide the QA for this SLR. The grading criteria are shown in Table III.

*Table III scoring criteria*

| SOURCES | RANKING | SCORE |
|---|---|---|
| Journal | Q1 | 2 |
|  | Q2 | 1.5 |
|  | Q3 OR Q4 | 1 |
|  | If the paper is not in JCR ranking | 0 |
| Conference | Core A | 1.5 |
|  | Core B | 1 |
|  | Core C | 0.5 |
|  | If paper is not in conference | 0 |

## IV. DATA ANALYSIS

This section reviews the outcomes and provides a comprehensive assessment of all of the papers. To properly address the research questions, the selected papers were studied. The first portion of this section goes over the search results that were retrieved using the specified search string. Following that comes a summary of the evaluation score, and the final section is devoted to in-depth discussions to address the research topics.

## A. SEARCH RESULTS

The utilization of data resources for the development of benchmark datasets, feature extraction, and computational models are only a few examples of the different factors that go into modern computer prediction of tomato late blight [30]. A total of 2270 items are produced by the primary search procedure from different web data sources. A screening procedure has been used focusing on the exclusion criteria, keywords, titles, abstracts, and full articles of retrieved publications, and a total of 240 papers were chosen as a result. Kappa coefficient is used for the selection of relevant papers. After that, we got rid of titles that were duplicates and titles that weren't pertinent to the evaluation. The selection procedure outlined in the preceding section was utilized in this collection. In addition, we have chosen 104 articles based on their abstracts after reading the entire abstracts of the 171 papers that were chosen during the duplicate phase. Out of 2700 investigations, 30 studies were ultimately chosen. The stages of the selection process are also shown in Figure 2 below, the outcomes of the selection and search operations are shown in table IV. and figure II presents the results of phase-by-phase selection.

*Table IV. The primary selection process for retrieved articles*

| Phase | Process | Selection Criteria | IEEE Xplore | Springer | Science Direct | Total |
|---|---|---|---|---|---|---|
| 1 | Search | Keywords | 1400 | 601 | 402 | 2270 |
| 2 | Screening | Title | 111 | 63 | 27 | 240 |
| 3 | Screening | Duplication | 72 | 53 | 29 | 171 |
| 4 | Screening | Abstract | 42 | 24 | 21 | 104 |
| 5 | Inspection | Full Article | 39 | 6 | 7 | 30 |

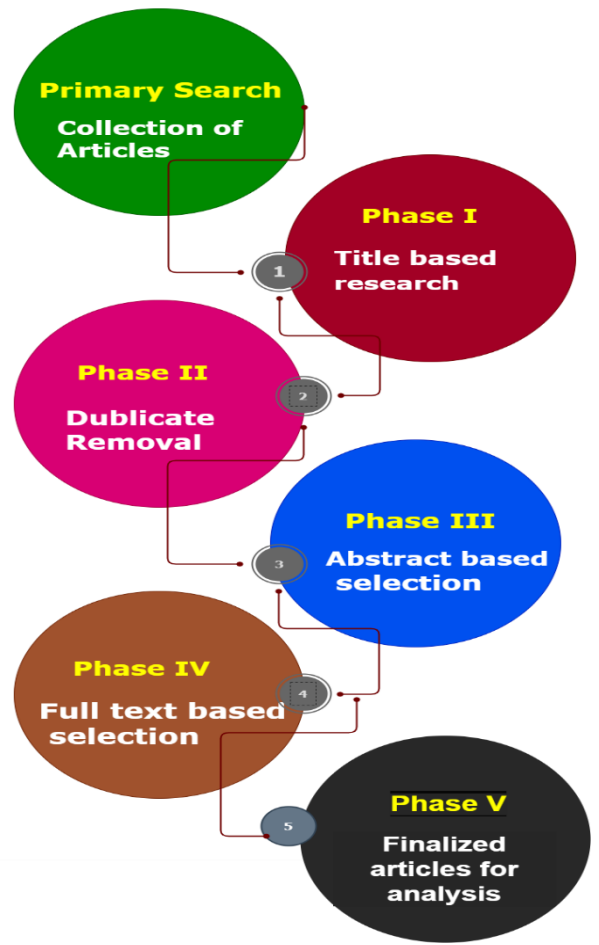

*Figure 2 phase-by-phase selection*

Figure 3 displays the distribution of selected papers and gives a graphic representation of all 55 finished articles, each of which has a unique publishing source. A symposium, a scientific report, and two more avenues of publishing were noted. At the conference, around 24% of the chosen papers were presented. whereas 57% of the research was presented at conferences and 2% in book chapters and symposiums. The distribution of the chosen papers are shown in FIGURE 4e.

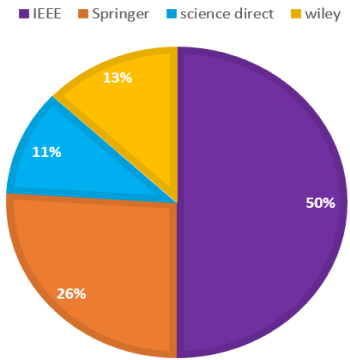

(a)selected conference papers distribution

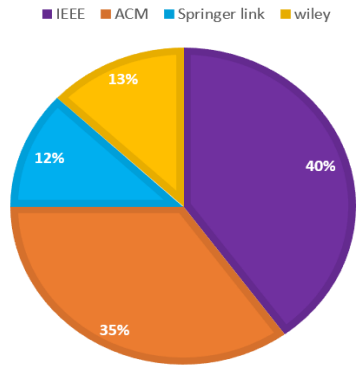

FIGURE 3. Selected papers distribution (a)selected conferences papers distribution (b) selected journal distribution

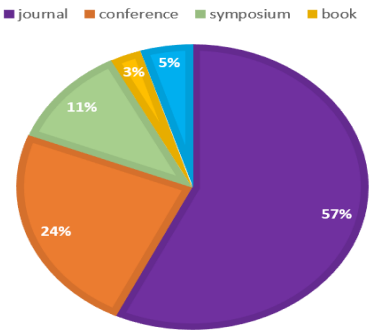

(b)selected journal papers distribution

Therefore, (a) Distribution of a few conference papers. (b) The distribution of a few journal articles. Additionally, Pie-Chart diagrams have been used to show the distribution of all sources, including IEEE, ACM, Springer, Science Direct, and Wiley, as well as the total number of selected articles from each source. In contrast, Fig. 4. (a) selected journal article distribution shows that overall papers from conferences have chosen IEEE, which has a share of 40%; similarly, springer has also share of 12%; ACM has a share of 35%, and Wiley has a share of 13%. Figure 4(b) shows the distribution of chosen journal papers. Springer received a 26 percent share of the papers, IEEE received a 50 percent part, Science Direct received an 11 percent share, and Wiley received a 13 percent share.

*Table v. Classification and quality assessment*

| Ref. No. | Year | Publication Type | Quality Assessment ||||||||  |
|---|---|---|---|---|---|---|---|---|---|---|---|
| | | | Major Focus Q1 | Major Focus Q2 | Major Focus Q3 | Criteria 1 | Criteria 2 | Criteria 3 | Criteria 4 | Criteria 4 | Total score |
| [1] | 2014 | Conference | Multi-class SVM. | Image celerity | Plant Village | 1 | 0 | 0.5 | 1 | 1 | 3.5 |
| [2] | 2018 | Conference | Image segmentation | visibility | Self-captured and Plant Village datasets | 1 | 1 | 1 | 1 | 1 | 5 |
| [3] | 2011 | Conference | Detection of late blight by morphological processing | Crop variety | Tomato field images | 1 | 1 | 1 | 1 | 1 | 5 |
| [4] | 2014 | Conference | Feature extraction | Multiple Disorder | Plant doc | 1 | 0.5 | 1 | 1 | 1 | 4.5 |
| [5] | 2021 | Journal | Image acquisition technique has been used | Leaf age | Mendeley dataset | 1 | 1 | 0.5 | 1 | 1 | 4.5 |
| [6] | 2017 | Journal | Detection of infected are using object detection techniques | dataset | Plant phenotyping dataset | 1 | 1 | 0.5 | 0.5 | 0.5 | 3.5 |
| [7] | 2019 | Conference | Automatic identification using image processing | segmentation | DiaMOS plant: | 0.5 | 0 | 1 | 1 | 1 | 3.5 |
| [8] | 2004 | Conference | Data Augmentation technique | Intrinsic factor | Flavia Dataset | 0.5 | 0.5 | 1 | 1 | 1 | 4 |
| [9] | 2010 | Journal | Residual deep CNN | Extrinsic | Plant Village | 0.5 | 0.5 | 1 | 0.5 | 1 | 3.5 |
| [10] | 2012 | Conference | Used K- means clustering | Segmentation issues | Plant doc | 0.5 | 0.5 | 1 | 1 | 1 | 4 |
| [11] | 2014 | Conference | Analysis on the use of applications smartphones for the diagnosis and monitoring of plant diseases. | Similar disorder | Plant Village (extended) | 1 | 1 | 0.5 | 1 | 1 | 4.5 |
| [12] | 2014 | Journal | Application of computational vision in agriculture to produce late blight | Less dataset | Plant Village | 0.5 | 1 | 1 | 1 | 1 | 4.5 |
| [13] | 2014 | Conference | frameworks in the literature for modeling disease detection | Controlled environment | PlantDoc | 0.5 | 0.5 | 1 | 1 | 0.5 | 2.5 |
| [14] | 2018 | Conference | Gradient Boosting Machine | empirical | alexnet | 0.5 | 1 | 1 | 1 | 1 | 4.5 |
| [15] | 2015 | Symposium | Image segmentation | Aged leaf | Tensorflow datasets | 1 | 0 | 1 | 1 | 1 | 4 |
| [16] | 2017 | Conference | LVQ Algorithm has been used | Illumination | Plant village | 1 | 0.5 | 1 | 1 | 1 | 4.5 |
| [17] | 2020 | Journal | hyperspectral remote sensing | noise | Plant doc | 1 | 0 | 1 | 1 | 1 | 4 |
| [18] | 2018 | Journal | a novel computer vision system has been proposed | No validation model | Mendeley dataset | 0.5 | 0.5 | 1 | 1 | 1 | 4 |
| [19] | 2012 | Journal | Classification technique has been used | Private datset | DiaMos Plant | 1 | 0 | 1 | 1 | 1 | 4 |
| [20] | 2019 | Journal | GLCM | Leaf growth | Plant village | 1 | 0.5 | 1 | 1 | 1 | 4.5 |
| [21] | 2021 | Journal | FUZZY Classification | Segmentation | Plant phenotyping dataset | 1 | 1 | 1 | 1 | 1 | 5 |
| [22] | 2018 | Conference | Feature extraction | Less work | TensorFlow | 1 | 1 | 1 | 1 | 1 | 5 |
| [23] | 2020 | Symposium | SURF algorithm is used | Camera issue | Flavia dataset | 1 | 1 | 0.5 | 0.5 | 1 | 4.5 |
| [24] | 2021 | Conference | ANN algorithm is used | Air factor | Plant village | 1 | 1 | 0.5 | 0.5 | 1 | 4 |
| [25] | 2018 | Journal | HSV transformation technique | Soil issue | PlantDoc | 1 | 1 | 1 | 0.5 | 1 | 4.5 |
| [26] | 2020 | Symposium | Hyperspectral remote sensing | Private data | Mendeley dataset | 1 | 1 | 1 | 1 | 1 | 5 |
| [27] | 2020 | Symposium | Back Propagation | less research | PlantDoc | 1 | 1 | 1 | 1 | 1 | 5 |
| [28] | 2020 | Conference | SGDM has been used | Generic factors | Plant Village (extended) | 1 | 1 | 1 | 1 | 1 | 5 |
| [29] | 2018 | Conference | SURF, RBF, SIFT has been used | Controlled | Plant Village (extended) | 0.5 | 1 | 1 | 0.5 | 1 | 4 |
| [30] | 2010 | journal | Nested–Leave–One–Out Cross validation | Extrinsic factors | .mendely data | 0.5 | 0.5 | 1 | 1 | 1 | 4 |

## B. ASSESSMENT AND DISCUSSION OF RESEARCH QUESTIONS

Based on the research topics listed in Table I, the key papers were examined in this section. The information obtained. After the examination of the chosen studies, problems regarding the evaluation of fragmented information were examined.

### 1) RQ1. WHAT TYPE OF IMAGE PROCESSING MODELS OR TECHNIQUES HAS BEEN USED TO IMPROVE THE QUALITY OF TOMATO CROPS AFFECTED WITH LATE BLIGHT DISEASE?

Using an integrated image processing approach, the developed technology recognizes tomato late blight leaf disorders. A filter is used to preprocess the collected photos. Multi threshold-based color segmentation is used to segment the preprocessed image. The spots are used to separate the images based on color. GLCM extracts the features of segmented images, which are subsequently picked using the Rough Set Approach. Maximum accuracy is attained using the KPCA 96.4.

### a). IMAGE PROCESSING

As illustrated in Fig.IV, the plant disease detection system procedure is divided into four parts. Gathering photos from a camera or a cell phone, and from the internet, is the first phase. The picture is divided into several clusters in the second stage, each of which can be treated differently], [12]. The Following figure shows the image processing techniques.

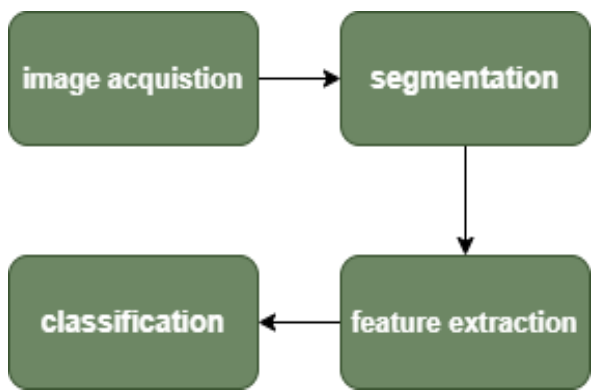

*Figure 4. Image Processing methods*

### b). IMAGE SEGMENTATION

By simplifying it, this stage aims to make an image's representation more understandable and scrutinizable. Because it serves as the basis for feature extraction, this stage of the process is also crucial to image processing. There are numerous methods for segmenting images, including thresholding, Otsu's algorithm [16], and k-means clustering. The k-means clustering algorithm divides objects or pixels into K groups established on a set of characteristics.

### C). TRAINING DATASET

For this research, a total of 106 digital photos of healthy and diseased late blight tomato plants were collected using a standard digital camera. from which 58 photos are healthy, and 48 images are affected with late blight [19]. The photos are utilized five times total at random for training and testing randomly selected photos from the whole dataset made up of 90%, 80%, 70%, 60%, and 50% were utilized for training. Randomly selected photos from the whole dataset were utilized for the testing, including 10%, 20%, 30%, 40%, and 50%. The suggested approach was carried out using MATLAB 2014b.

### d). PREPROCESSING:

Apply contrast restricted adaptive histogram equalization to RGB photos of late blight and healthy areas to improve the contrast of the intensity.

### e) FEATURE EXTRACTION

Wavelet characteristics are extracted from pictures of tomato late blight-affected and healthy plants using wavelet transformation. As wavelet transformation divides the digital signal into wavelet functions, it converts the digital signal to a wavelet signal [13]. The wavelet transformation can analyze the vertical, diagonal, and horizontal picture sub bands in depth. By using the ow pass and high pass filtering, the detailed information of the picture may be extracted [14]. The experiment's discrete wavelet transform had an impact on the plant's area when characteristics from the illness were extracted. R, G, and B components were first separated independently from the photos. Using a single level, two-dimensional discrete Haar wavelet, calculate the low and high frequencies of each R, G, and B component.

### f). EVALUATION

Euclidean measured the distance between the training set and the testing set using the distance technique [15] An unknown item, also known as a test object, can be recognized by connecting it with one of those classes using similarity (or, alternatively, dissimilarity or distance) measurements whether through supervised learning or unsupervised learning are known. To compare or distance between two groups, such as a tomato with late blight or a healthy picture. Figure IV shows the Steps for Recognition and Classification of Tomato Late Blight by using DWT and Component Analysis

### g). RESULTS AND DISCUSSION

Accuracy of 96.4 percent was achieved in the identification and categorization of tomato late blight utilizing discrete wavelet transformation and component analysis. Utilizing wavelet features, the PCA, KPCA, and ICA are utilized to reduce the dimensionality of the data before being used to identify and categorize tomato plant diseases. To determine if a picture was impacted by the late blight illness or not, the two classes—late blight infected and healthy—were employed [25]. The Euclidean distance metric was used to categorize whether the testing picture belongs to which class—infected or healthy—by computing the smallest distance between training and testing data. Table VI shows the recognition and classification of late blight accuracy

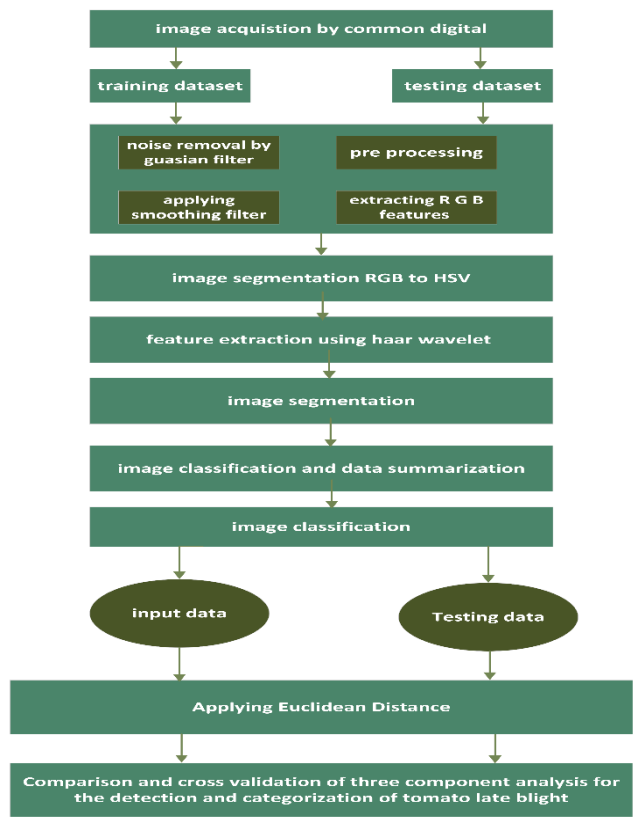

*Figure 5. Steps for Recognition and Classification of Tomato Late Blight by using DWT and Component Analysis*

TABLE VI. RESULTS

|  | PRINCIPAL COMPONENT ANALYSIS | | KERNAL PRINCIPAL COMPONENT ANALYSIS | | INDEPENDENT COMPONENT ANALYSIS | |
|---|---|---|---|---|---|---|
| TRAINING DATASET | HEALTHY | LATE BLIGHT INFECTED | HEALTHY | LATE BLIGHT INFECTED | HEALTHY | LATE BLIGHT INFECTED |
| 10% | 100% | 100% | 100% | 0% | 100% | 24% |
| 20% | 89% | 90% | 90% | 75% | 25% | 55% |
| 30% | 90% | 93% | 95% | 10% | 100% | 67% |
| 40% | 87% | 89% | 48% | 47% | 65% | 44% |
| 50% | 97.9% | 96% | 20% | 34% | 55% | 76% |
| Overall Accuracy | 90% | 75.8% | 61.2% | 28% | 69% | 51.1% |

2) *RQ 2. WHAT TYPE OF CHALLENGES AND OPPORTUNITIES EXISTING IMAGE PROCESSING FOR THE DIAGNOSIS OF LATE BLIGHT?*

**CHALLENGES AND OPPORTUNITIES**

The use of image processing techniques for the identification of late blight presents both many potentials and limitations. The problems that were noted in the literature are discussed in this section.

**A. DATASETS VARIATION**
For the categorization of late blight, many datasets were provided. While some datasets were accessible to the public, others weren't. Different datasets were found to have varying quantities of photos. Additionally, several authors created their own picture datasets by using the internet [21].

**1) LIMITED NUMBER OF IMAGES IN DATASETS**

There aren't many photos in the benchmark datasets that are now available for training and testing, and it has also been noted that there aren't many images in the benchmark datasets that can be used for both training and testing. The reliability of the proposed approaches remains unknown when tested on a big picture collection, despite their good performance on a limited number of photos [17]. Only 200 photos are included in the PH2 training and testing dataset. ISIC announces an annual challenge to address the outlined to resolve this issue. ISIC offers a yearly competition to address the identified issue to remedy it. Additionally, some researchers merge several datasets to create a single, sizable picture dataset, which they then use to test their suggested approaches [18].

**2) NON-PUBLIC DATASETS**
Some researchers make use of photos gathered from the internet and private databases. Therefore, because of the dataset's unavailability, makes the task of replicating the results more difficult [19].

**B. SIZE OF SPOT**
Additionally, it is shown that the spot's dimension plays a crucial role [20]. Late blight is tricky to identify, and detection. accuracy drops significantly if the lesion is less than 5mm in size.

**C. POTENTIAL CLASSIFIERS FOR LATE BLIGHT**
Pre-trained models and manually created deep learning approaches have demonstrated promising outcomes for late blight diagnosis with high accuracy, according to the research. Pre-trained models and handmade methods established on image processing were shown to provide promising outcomes for late blight recognition with high accuracies in the literature [22].

**D. ACCURACY OF IMAGE PROCESSING METHOD**
According to the research, pre-trained models and manually developed deep learning algorithms have shown favorable results for late blight recognition with high accuracy [23].
In the literature, it has been demonstrated that deep learning-based handcrafted approaches and pre-trained models may diagnose late blight with excellent accuracy. Figure x shows the challenges.

**E. ILLUMINATION CONDITIONS**
The effectiveness of the approaches can also be impacted by the amount of light used to capture the images. Brightness may be varied and is considerably diffused in lab lighting. Nevertheless, in suitable field circumstances, lighting may differ and associated attributes may degrade, which may impair the algorithms' ability to provide accurate answers[12][26]. Additionally, there may be fluctuations in the amount of sunshine and intensity as the day progresses and as it comes to a close. Regardless of whether rectification or correction is made to the photos, many of the qualities that are often extracted from photographs are susceptible to specific variances. [27].

**F. IMAGE BACKGROUND**
Leaf segmentation is the starting point for almost all methods used for leaf analysis in picture classification. By keeping a panel (blue or white) underneath the leaf, this step may be made automatically and extremely easy. On the other hand, several additional context-related features, such as the soil, plants, and leaves, might make segmentation challenging. It's particularly challenging to split the leaf when the background has a lot of green features. As long as the backdrop doesn't include leaves that are similar to the one selected for study, segmentation can be successfully accomplished by reducing the utilization of visible light. A multispectral camera and all its filters were used by one researcher [31] to study this and distinguish the backdrop made up of tulip plants. Figure 6 makes it very clear that the results will be hampered by the image's noise. Using the most recent edge detection techniques, which can greatly enhance segmentation, the best results can be discovered in images for the delimitation of all the elements that are currently present. Even though during picture capture, the image's features are crisp and concentrated in compared to the rest of the image, a better technique for identifying and separating the objects of interest requires the establishment of a metric for the sharpness of those objects' main features.

**G. SENSOR DIFFERENCS**

Technical elements including the kind of optical property, dynamic range, and pixel sensor vary greatly different distribution models, producing pictures with a variety of characteristics. Considering (atmospheric factors) such humidity, light exposure, and temperature, Even though the camera used to take the photos is the same, its attributes might change. [36]. Additionally, several picture qualities are substantially influenced by camera attributes including image stabilization, ISO sensitivity, image compression, aperture, and shutter speed [37], while the impacts of these attributes may be adjusted with the use of image calibration. It's significant to note that several studies don't include enough information about the experimental design and the kinds of sensors utilized to permit reliable replication of the study.

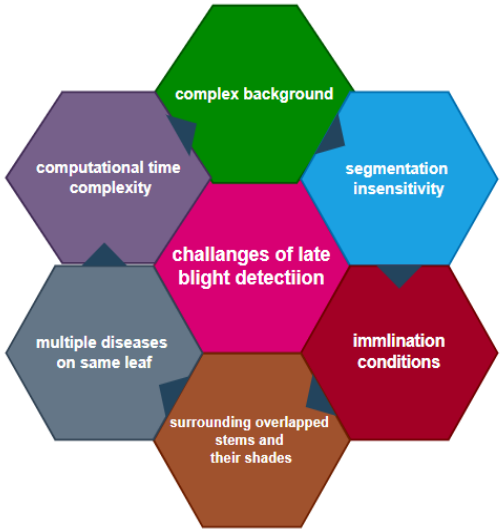

*Figure 6 challenges in detection of late blight*

### 3) RQ 3. WHAT TYPE OF DATASETS ARE AVAILABLE FOR THE DETECTION OF LATE BLIGHT?

The identification of late blight was possible using a variety of data sets. Some of them were accessible to everyone, while others weren't. Due to their extensive use in research for the identification of late blight, the datasets mentioned below were taken into consideration as a benchmark. Table VII. provides a summary of the benchmark datasets.

**a) ISBI Challenge 2019 Dataset**: The research articles have used this dataset to assess the effectiveness of the methods they have advised. The dataset challenge includes 379 evaluation pictures and 273 training images (115).

**b) PlantDoc:**
These datasets are used in this review by articles. For the diagnosis of late blight disease [24], plant doc offers photos. Plantdoc is the most comprehensive picture database for diagnosing plant diseases online. It has 150 pictures of late blight. [26]

**c) Mendeley Data**:
According to several picture sources, his database is split into two datasets for tomato leaf photos. The first dataset's tomato leaf photos were chosen from a database called Plant Village which has ten categories (nine disease categories and one health). There are a total of 14,531 photos [27], each of which consists of a single leaf and a single backdrop. The picture size was changed from 256 X 256 to 227 * 22 after unifying the original tomato leaf photos and eliminating superfluous categories [30]

**d) Plant phenotyping dataset:**
It consists of 1000 tomato leave diseases out of which 70 were of late blight disease.

**e) DiaMOS plant**:
set of 23,906 publicly accessible photos in an image dataset. This dataset is used in this review by [65], [68], [70], [74], and [92] to assess the efficacy of their techniques.

**f) Flavia Dataset [30]:**
For the purpose of testing their suggested approaches, [46], [49], [56], [57], [61], [62], [69], and [75] utilized this dataset. 374 of the 2,000 photos totaling 2,000 were late blights.

Table *VII the recognition and classification of late blight accuracy* VII

| Dataset | No.of images | Late blight images |
|---|---|---|
| ISBI Challenge 2019 Dataset | 379 | 115 |
| PlantDoc | 1500 | 150 |
| Mendeley Data | 14531 | 50 |
| Plant phenotyping dataset | 1000 | 70 |
| DiaMOS plant | 23906 | 125 |
| Flavia Dataset | 2000 | 374 |

## VI. DISCUSSION

The SLR gathered the research on the detection of late blight disease by image processing methods. The research questions set at the start of this study have been addressed by these studies. The findings indicate that additional solutions have been put forth in an effort to identify practical remedies to the issues with late blight disease detection [28].

### A. PROPOSED TAXONOMY
Many methods for the identification of plant diseases have been presented in recent years. These methods weren't evaluated or validated for their efficacy or accuracy before being offered. Additionally, the assessment and validation studies that were created did not evaluate the methods for disease detection using standard quality criteria. In other words, these methods examined several elements.

Consequently, the reviewed research do not agree on a theory in which particular. A detection method developed in respect to the dimensions involved. The works under consideration show the necessity of doing applied research, i.e., validating the models put forward in the literature. In this way, promoting the outcomes using software platforms linked to these quality indicators to show the precision and efficacy of the methods employed. To address this, a quality model for assessing methods for identifying plant diseases in empirical investigations is proposed. Studies that apply technologies to the field of agriculture research generally lack a good model. In particular, we were unable to find any experimental study that evaluated these methods for sickness diagnosis using a shared quality model. Therefore, it is important to assess the quality of plant disease detection systems by focusing on metrics like precision, accuracy, recall, and f-measure. As a result, developing a quality model would be a fascinating study subject that the scientific community may examine in future studies.

Three stages are established for the evaluation and validation process; the first stage compares the approach to other state-of-the-art methods to determine how effective each is at segmenting the background. In the third stage, this process is verified by identifying additional illnesses that, in terms of color, resemble Phytophthora infectants. In this instance, three state-of-the-art methods are used for comparison as they don't use the image's backdrop. The Plant Village database, which contains pictures of all tomato leaves, both healthy and sick, with a segmented backdrop.

The G, B, and S channels from the RGB and HSI are employed in the third method. Finally, using both black and yellow sigatoka, this process was tried on banana leaves. When this disease first appears, it causes 2-4 leaves to have vertical stripes of red brown color. As the lesion worsens, the stripes turn dark brown and develop a yellow halo. This disease is known as yellow Sigatoka, and it is very prevalent in cool climates and elevations higher than 1600 meters above sea level. Vertical yellow streaks are the early sign of this illness. As the lesion develops, the color intensifies, making it more noticeable, and eventually turns brown with a black border. Given that these two illnesses' basic traits like those of Phytophthora infectants, it was chosen to use the approach suggested in It required the creation of a database with 465 photographs. Therefore, real-time categorization systems (using cameras to monitor the plantations) are required, significantly increasing the amount of data that has to be analyzed.

In small rural estates, it is challenging to develop low-cost machine learning models for predicting future microclimates. The three necessary variables—precipitation (pluviometer), temperature (thermometer), and humidity (hygrometer)—indicate the development of diseases (fungi), and some of these variables have consistent, predictable values.

Therefore, we draw the conclusion that in order to get values that are more in line with what has been accomplished, it will be essential to use several approaches on each label. This data collection could need the use of numerous techniques rather than just one. For instance, the label "precipitation," which refers to the occurrence of rain, does not demonstrate consistency in its values and does not provide confident forecasts.

A precise alignment is required for the fungus to grow in the plant and is dependent on the set of environmental parameters (labels) supplied. Anticipating the use of pesticides is extremely important, as it is difficult to develop microclimate predictions of up to 12 hours. To determine the precise patterns that enable it to recognise the disease and its mutations throughout time, a network of a specific fungal disease will be trained using meteorological data from the future.

The second approach deals with classification and creates strong models that can be included into high-throughput systems by using field-condition picture data (i.e., data with variations in lighting and shading) for the generation of training data sets.

Taxonomy below represents the summary of the area of study examined. Seven categories make up the taxonomy: (1) encouraged Six categories for detection and prediction were
found using technology of plant ailments; (2) types of sensors, including those utilized for data collection were split into two groups. Many sensors, including image sensors; (3) 11 kinds of architectures examined the use of neural networks and hierarchical representations; (4) contribution; (5)growth factors of late blight; (6) infected areas and color and shapes

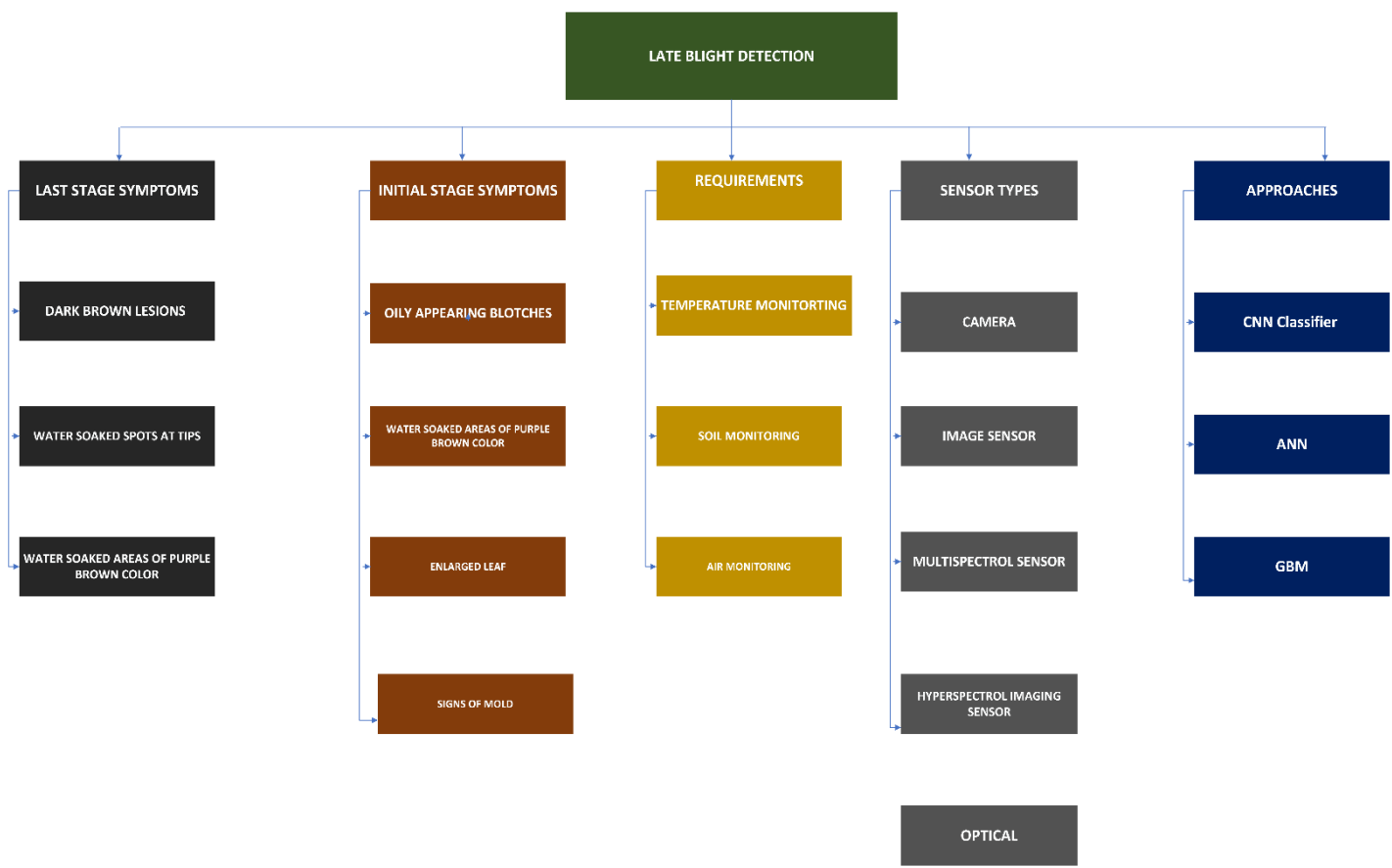

*Figure 7 taxonomy of late blight*

## B. PROPOSED MODEL

The key step of the proposed model is represented in Fig. vii. Firstly, hyperspectral images of healthy and sick tomato leaves were produced using the hyperspectral imaging equipment in the spectral wavelength range of 380–1023nm. The reflectance values of every pixel in the corrected hyperspectral images' region of interest (ROI) were retrieved and used as X variables. Every sample was split into training and testing sets in a 2:1 ratio. The conveyer belt was used to position each tomato leaf for line-by-line scanning with hyperspectral imaging equipment. The movement speed was 3.3 mm/s and the exposure duration was 0.065 s. The tomato leaf was 38.0 cm away from the camera in the vertical direction. When the tomato leaf was scanned, a hyperspectral picture was produced by employing the imaging spectrograph in the spectral range of 380 to 1023nm. The spatial dimension of the hyperspectral cube comprised 672 pixels, while the spectral dimension had 512 bands. Using black-and-white reference photos, raw hyperspectral images were adjusted into the reference images extraction of spectral information. A ROI with 2020 pixels was carefully chosen from each leaf to lessen the impact of sample inequity and pixel variation. The ROIs for sick leaves were found in the symptomatic regions. And the ROIs were located where they should have been for healthy samples. Next, the average spectral reflectance values of each ROI's pixels were calculated. As a result, there were 103 samples for the testing set and 207 samples for the training set (the numbers for healthy, early blight, and late blight samples are, respectively, 80, 80, and 47.

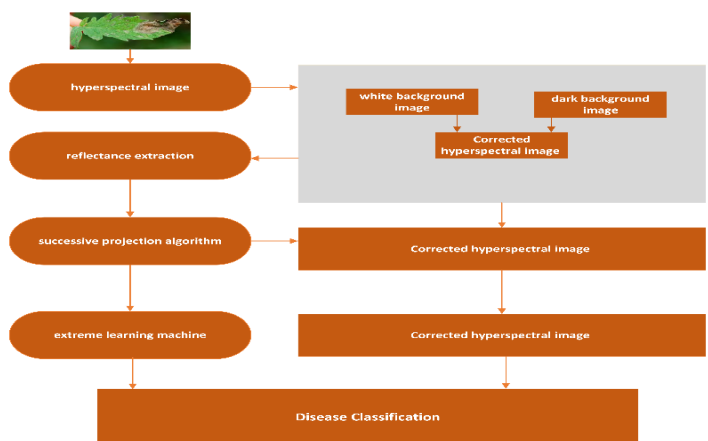

**Figure 8 late blight detection model**

**C. OPEN ISSUES AND RESEARCH GAPS**

- A high-quality model for assessing methods for identifying and categorizing plant diseases. Many methods for the identification of plant diseases have been presented in recent years. 18% of the primary papers suggested methods for detecting late blight disease. These methods weren't validated or evaluated for their efficacy and accuracy before being proposed. Research on how to identify plant diseases empirically.

- Empirical research makes up the majority of primary studies (59%, 33/56). Many evaluation (34%, 19/56) and validation (25%, 14/56) studies were produced, in particular. This demonstrates that the agricultural research community is concerned in using computing technologies to combat disease. Empirical studies make up 59% of primary studies (of 56 total). Many evaluation (34%, 19/56) and validation (25%, 14/56) studies were produced, in particular. This demonstrates that the agricultural research community has been interested in producing empirical studies to help combat the illness. Even Nevertheless, research has not yet adequately addressed all aspects of illness detection, thus future research should continue to place a strong emphasis on performing empirical investigations to assess the efficacy of detection methods.

- Disorders with similar symptoms
Many researchers adopt the implicit assumption that some nutrient-deficient plants are caused by only one or a few circumstances based on their spectral pictures and visual symptoms.

- Segmentation in plants
One unexpected effect of such an approach might dissuade many users, who may be the potential users, from using the techniques because of the significant additional work necessary to fulfil these requirements. Lack of distinct boundaries, noisy backgrounds, diseased symptoms, variation present due to the environment during image capture, the same diseases causing different symptoms, multiple disorders producing the same symptoms simultaneously, and symptoms produced by multiple disorders producing the same symptoms continue to play a critical part directly affecting the effectiveness of the image processing methods that have been proposed.

**D. FUTURE WORK**

Future study could go in many different areas. To get more accurate results, one way is to increase the dataset. Finding a solution to the issue of detection failures brought on by poor image resolutions is another avenue. The strategy could also be applied to other crops in addition to tomatoes. Using the occurrence of several micro-climates in the same environment as a starting point, it is challenging to predict the limiting factors to climate variables in the context of new developing technologies. Therefore, the development and use of a system to identify the specific illness stage will be of tremendous importance. Early disease detection, sometimes referred to as disease prediction, enables farmers to take the appropriate safeguards and thus lower the percentage of crop damage. Therefore, I suggest that the implementation in practice be done in a subsequent effort through the creation of a real-time system. identifying plant diseases in a field setting by using computer vision methods, edge computing, continuous monitoring, and prediction algorithms. Finally, it is hoped that the findings presented in this work will inspire scholars and professionals to further investigate the findings. This research may also be considered as the first stage in a more ambitious effort to describe and enhance methods for plant disease detection and prediction in order to better safeguard crops and minimize financial losses for small farmers and investors, particularly those with limited resources.

**VII. CONCLUSION:**

Modern research for detecting late blight has been addressed in this study. Moreover. Open problems and difficulties have been noted. Furthermore, various image processing methods to find late blight It has been noted that applying image processing techniques eliminates the need for elaborate and composite pre-processing methods including pixel value normalization, cropping, and picture resizing [30]. By reviewing pertinent literature, a suggested taxonomy and a proposed model have been provided. Additionally, this analysis lists the primary flaws in the current approaches and highlights the areas that require improvement. For researchers to assess their work, a substantial number of labeled benchmark datasets have been made available, including DiaMos, ISBI (2016, 2017, and 2018 challenges), plan doc, and open-access datasets. Additionally, datasets that had not been published or listed were also accessible for the identification of late blight. However, the availability of so many different datasets make it challenging to compare the results. Future study must make use of a bigger dataset and fine-tune hyper-parameters to lower the likelihood of overfitting. Maximum accuracy is attained with KPCA of 96.4. The issue of raising the accuracy rate still exists, though. Maximum sensitivity must be attained while also enhancing the approaches' specificity and overall accuracy.

**Table VII Publication score**

| Publication score | channel | reference | No. | % |
|---|---|---|---|---|
| 2016 IEEE conference on image processing | Conference | [1] | 1 | 1.6 |
| 2017 international Journal of Advanced Research in Computer and Communication | Conference | [2] | 1 | 1.6 |
| Proceedings of the SMART–2019, IEEE Conference International Conference on System Modeling & Advancement in Research Trends | Conference | [3] | 1 | 2.7 |
| IEEE Systematic review of image processing techniques in plant disease detection 2020 | Journal | [4] | 2 | 4.7 |
| (IJACSA) International Journal of Advanced Computer Science and Applications, Vol. 13, No. 1, 2022 | Journal | [5] | 1 | 1.6 |
| 1st KEC Conference Proceeding 2018 | Conference | [6] | 1 | 1.6 |
| Plant diseases recognition on images using convolutional neural networks: A systematic review | Journal | [7] | 2 | 1.6 |
| Technological support for detection and prediction of plant diseases: A systematic mapping study | Conference | [8] | 1 | 1.6 |
| I.J. Engineering and Manufacturing, 2015, 4, 12-22 | Journal | [9] | 3 | 2.3 |
| 2021 2nd International Conference on Intelligent Engineering and Management (ICIEM) | Conference | [10] | 3 | 3.1 |
| International Conference on Communication and Signal Processing, July 28 - 30, 2020 | Conference | [11] | 1 | 1.6 |
| 2019 5th International Conference on Advanced Computing & Communication Systems (ICACCS) | Conference | [12] | 1 | 1.6 |
| Disease Detection on the Leaves of the Tomato Plants by Using Deep Learning | Journal | [13] | 1 | 1.6 |
| Identification of Tomato Disease Types and Detection of Infected Areas Based on Deep Convolutional Neural Networks and Object Detection Techniques 2019 | Journal | [14] | 3 | 1.6 |
| IEEE Journal of Agricultural Sciences 90 (2): 249–57, February 2020/Review Article | Journal | [15] | 3 | 1.6 |
| Plant Disease Detection Using Image Processing and Machine Learning | Journal | [16] | 2 | 1.6 |
| PLANT DISEASES RECOGNITION ON IMAGES USING CONVOLUTIONAL NEURAL NETWORKS: A SYSTEMATIC REVIEW | Journal | [17] | 1 | 2.4 |
| VFAST Transactions on Software Engineering 2411-6246 Volume 9, Number 4, October-December 2021 | Journal | [18] | 1 | 1.6 |
| International Journal of Advanced Research in Electrical, Electronics, and Instrumentation. Volume 7, Issue 12, December 2018 | Journal | [19] | 1 | 1.6 |
| International Journal of Applied Earth Observation and Geoinformation 4 (2003) 295–310 | Journal | [20] | 1 | 1.6 |
| 8th International Conference on System Modeling & Advancement in Research Trends, 22nd–23rd November, 2019 | Conference | [21] | 1 | 1.6 |
| 2018 IEEE 4th International Symposium in Robotics and Manufacturing Automation (ROMA) | symposium | [22] | 1 | 4.7 |
| Proceedings of the International Conference on Intelligent Sustainable Systems (ICISS 2017) IEEE Xplore | Conference | [23] | 1 | 1.6 |
| 2016 Online International Conference on Green Engineering and Technologies (IC-GET) | Conference | [24] | 2 | 1.6 |
| International Journal of Engineering and Advanced Technology (IJEAT) ISSN: 2249 – 8958, Volume-8, Issue-3S, February 2019 | Journal | [25] | 1 | 1.6 |
| International Journal of Computer Applications (0975 – 8887) | journal | [26] | 1 | 1.6 |
| 2019 International Conference on Automation, Computational and Technology Management (ICACTM) Amity University | Conference | [27] | 2 | 2.4 |
| 2020 International Conference on Contemporary Computing and Applications (IC3A) | Conference | [28] | 1 | 1.6 |
| Journal of Scientific & Industrial Research Vol. 79, May 2020, pp. 372-376 | Journal | [29] | 1 | 1.6 |
| International Journal of Computer Engineering and Applications, Volume VI, Issue-III June 2014 | Journal | [30] | 3 | 1.4 |

**References**

[1] S.K.R. Yellareddygari, R.J. Taylor, J.S. Pasche, A. Zhang, N.C. Gudmestad, Predicting potato tuber yield loss due to early blight severity in the Midwestern United States, Eur. J. Plant Pathol. 152 (1) (2018) 71–79.

[2] Vanhaelewyn L, Van Der Straeten D, De Coninck B and Vandenbussche F (2020) Ultraviolet Radiation From a Plant Perspective: The Plant-Microorganism Context. Front. Plant Sci. 11:597642.

[3] Y. Tang, S. Dananjayan, C. Hou, Q. Guo, S. Luo, Y. He, A survey on the 5G network and its impact on agriculture: challenges and opportunities, Comput. Electron. Agric. 180 (2021), 105895.

[4] Rathod Arti N, Tanawal Bhavesh, Shah Vatsal. Image processing techniques for detection of leaf disease. Int J Adv Res Comput Sci Softw Eng 2013; Page-3.

[5] Shivani K. Tichkule, Dhanashri. H. Gawali. Plant diseases detection using image processing techniques. Conference: 2016 Online International Conference on Green Engineering and Technologies (IC-GET).

[6] C. C. Tucker and S. Chakraborty, "Quantitative assessment of lesion characteristics and disease severity using digital image processing," *Journal Phytopathology*, vol. 145, pp. 273–278, 1997.

[7] Minghua Zhang, Zhihao Qin, Xue Liu. Remote sensed spectral imagery to detect late blight in field tomatoes. Precision Agriculture, Springer 2005;6:489–508.

[8] S. Arivazghan, R. Newlin Shebiah, S. Ananthi, S. Vishnu Varthini. Detection of Unhealthy Region of Plant leaves and Classification of Plant Diseases using Texture Features. Agricultural Engineering International: CIGR Journal 2013;15(1):211-217.

[9] Niket Amoda, Bharat Jadhav, Smeeta Naikwadi. Detection and Classification of Plant Diseases using Image Processing. International Journal of Innovative Science, Engineering and Technology 2014;1(2): 1- 7.

[10] Guili Xu, Fenling Zhang, Syed Ghafoor Shah, Yongqiang Ye, Hanping Mao. Use of leaf color images to identify nitrogen and potassium deficient tomatoes. Pattern Recognition Letters 2011;32(11):1584-1590.

[11] Jagdish, S.V.K, Murty, M.V.R, Quick, W.P., Rice responses to rising temp

– challenges, perspectives and future directions, *Plant, Cell and environment, vol. 38, Issue. 9,* pp: 1686-1698, 2015.

[12] Liu and Wang Plant Methods (2021) 17:22. Plant diseases and pests detection based on deep learning: a review

[13] John W. Woods, Wavelet Theory. in Multidimensional Signal, Image, and Video Processing and Coding (Second Edition), 2012

[14] https://www.l3harrisgeospatial.com/docs/highpassfilter.html

[15] BROWNLEE, J. 4 DISTANCE MEASURES FOR MACHINE LEARNING. RETRIEVED FROM HTTPS://MACHINELEARNINGMASTERY.COM/DISTANCE-MEASURES-FOR-MACHINE-LEARNING

[16] NORHANIS AYUNIE AHMAD KHAIRUDIN ET AL 2020 . IMAGE SEGMENTATION USING K-MEANS CLUSTERING AND OTSU'S THRESHOLDING WITH CLASSIFICATION METHOD FOR HUMAN INTESTINAL PARASITES. IOP CONF. SER.: MATER. SCI. ENG. 864 012132

[17] NIGHTINGALE, S.J., WADE, K.A. & WATSON, D.G. CAN PEOPLE IDENTIFY ORIGINAL AND MANIPULATED PHOTOS OF REAL-WORLD SCENES?. *COGN. RESEARCH* 2, 30 (2017). HTTPS://DOI.ORG/10.1186/S41235-017-0067-2

[18] Kotu, V. & Deshpande, B. (2019). Data science, Concepts and practice. Elsevier. Morgan Kaufmann.

[19] Lynette M. Smith, Fang Yu & Kendra K. Schmid (2021) Role of Replication Research in Biostatistics Graduate Education, Journal of Statistics and Data Science Education, 29:1, 95-104, DOI: 10.1080/10691898.2020.1844105

[20] Khoury, B.M., et. al. (2015) The Use of NanoComputed Tomography to Enhance Musculoskeletal Research. Author manuscript.

[21] 50 FREE MACHINE LEARNING DATASETS: IMAGE DATASETS. RETRIEVED FROM HTTPS://BLOG.CAMBRIDGESPARK.COM/50-FREE-MACHINE-LEARNING-DATASETS-IMAGE-DATASETS-241852B03B49

[22] LAWRENCE C. NGUGI, MOATAZ ABELWAHAB, MOHAMMED ABO-ZAHHAD (2021)ECENT ADVANCES IN IMAGE PROCESSING TECHNIQUES FOR AUTOMATED LEAF PEST AND DISEASE RECOGNITION – A REVIEW .*Information Processing in Agriculture*. 8(1)27-51.

[23] Tilahun, N. & Assefa, B.G (2020). Artificial Intelligence Assisted Early Blight and Late Blight Potato Disease Detection Using Convolutional Neural Networks. Conference: 18th Biannual Conference of Crop Science Society At: Addis Ababa Volume: 8

[24] AMANDA GEVENS, ANNA SEIDL AND BRIAN HUDELSON. LATE BLIGHT. RETRIEVED FROM HTTPS://PDDC.WISC.EDU/2015/07/28/LATE-BLIGHT/

[25] Xie C, Shao Y, Li X, He Y. Detection of early blight and late blight diseases on tomato leaves using hyperspectral imaging. Sci Rep. 2015 Nov 17;5:16564. doi: 10.1038/srep16564. PMID: 26572857; PMCID: PMC4647840.

[26] HASAN, REEM IBRAHIM, YUSUF, SUHAILA MOHD, & ALZUBAIDI, LAITH (2020) REVIEW OF THE STATE OF THE ART OF DEEP LEARNING FOR PLANT DISEASES: A BROAD ANALYSIS AND DISCUSSION. PLANTS, 9(10), ARTICLE NUMBER: 1302.

[27] Bhujel, A.; Kim, N.-E.;Arulmozhi, E.; Basak, J.K.;Kim, H.-T. A Lightweight Attention-Based Convolutional NeuralNetworks for Tomato Leaf DiseaseClassification. Agriculture 2022, 12,228. https://doi.org/10.3390/agriculture12020228

[28] Lal, Mehi et al. "Management of Late Blight of Potato". Potato - From Incas to All Over the World, edited by Mustafa Yildiz, IntechOpen, 2018. 10.5772/intechopen.72472.

[29] Newlands NK (2018) Model-Based Forecasting of Agricultural Crop Disease Risk at the Regional Scale, Integrating Airborne Inoculum, Environmental, and Satellite-Based Monitoring Data. Front. Environ. Sci. 6:63

[30] AAKANKSHA NS (2020). IMAGE PROCESSING AND DATA AUGMENTATION TECHNIQUES FOR COMPUTER VISION. RETRIEVED FROM: HTTPS://TOWARDSDATASCIENCE.COM/IMAGE-PROCESSING-TECHNIQUES-FOR-COMPUTER-VISION-11F92F511E21